\documentclass{svproc}
\usepackage{url}

\usepackage{graphicx}
\usepackage{float}
\usepackage{amsmath}
\usepackage{booktabs}
\usepackage{xcolor}

\begin{document}
\mainmatter              
\title{Improving Quantized Model Performance in Qualitative Analysis with Multi-Pass Prompt Verification}
\titlerunning{Contribution Title}  
%
\author{Aisvarya Adeseye\inst{1} \and Jouni Isoaho\inst{2} \and Adeyemi Adeseye\inst{3}}
\authorrunning{A. Adeseye et al} 
%
%
\institute{University of Turku, Turku, Finland,\\
\email{aisvarya.a.adeseye@utu.fi}\\ 
\and
University of Turku, Turku, Finland\\
\email{jouni.isoaho@utu.fi} \\
\and
Brilloconnetz Partners avoin yhtiö, Turku, Finland\\
\email{adeyemi@brilloconnetz.com}}

\maketitle              

\begin{abstract}
Quantized Large Language Models (LLMs) are used more often in qualitative analysis because they run fast and need fewer computing resources. This study examines how different lower bits quantization levels (8-bit, 4-bit, 3-bit, and 2-bit) and quantization types affect the performance of LLaMA-3.1 (8B) on qualitative analysis. The study uses expert and non-expert responses from 82 interview transcripts. Low-bit models often produce higher levels of hallucinations and unstable results, especially when reading non-expert language with unclear terms.
To improve performance, we propose a quantization-aware multi-pass prompt verification method. This method guides the model through controlled steps that reduce hallucinations. It removes unreliable content and passes the results to the next transcript after verification, improving accuracy.
To validate performance, human coders analyzed transcripts using NVivo and BF16 LLaMA. BF16 LLaMA-3.1 produced high-precision output but had semantic drift and hallucination. These errors were corrected manually. The corrected BF16 output and NVivo human coding were combined to create a gold-standard ground truth (GSGT) for thematic extraction and frequency analysis.
The results show that 8-bit models stay closest to the GSGT. The 4-bit models lose accuracy but become stable when the proposed method is applied. The 3-bit and 2-bit models drop in performance because of heavy compression, but they improve with the proposed prompt design and verification. The study also finds that models at the same bit level behave differently depending on quantization type. Overall, the method helps low-resource LLMs become more stable, accurate, and suitable for qualitative research at lower cost.
\end{abstract}
\section{Introduction}

Large Language Models (LLMs) are increasingly used in qualitative analysis because they can summarize long texts, identify themes, and support coding tasks with high speed and low manual effort \cite{adeseye2025llmqual}. These models help researchers process large datasets, reduce human workload, and support reflective analysis in fields such as social science, cybersecurity, behavioral studies, and healthcare research \cite{bano2023llmqual,rossi2024llmdata}. However, most existing work relies on full-precision LLMs or commercial API-based systems \cite{fischer2024exploring,castellanosreyes2025transforming}. These systems raise two major challenges: high computational cost and privacy risks \cite{adeseye2025efficientprompt}. 
Quantization offers a promising solution as it helps reduce model size and consequently resource usage. It lowers the bit precision of model weights and activations, allowing LLMs to run on smaller hardware with lower cost \cite{frantar2023gptq}. However, quantization also increases hallucination, semantic drift, and loss of accuracy, especially when reading non-expert language, which often contains unclear or inconsistent wording. Prior studies have explored hallucination reduction \cite{bai2025hallucination,adeseye2026hallucination} and prompt design \cite{li2024selfprompting}, but they do not examine how different quantization levels and quantization types affect LLM performance in qualitative analysis. They also do not explore how expert and non-expert terminology changes the behavior of quantized LLMs. Additionally, previous work rarely compares quantized LLM outputs against strong human-coded benchmarks using tools such as NVivo.


The aim of this study is to develop and evaluate a quantization-aware multi-pass prompt verification framework to improve the accuracy, reliability, and hallucination result of quantized LLaMA-3.1 (8B) models for qualitative interview analysis.
The study has three main objectives:  
\begin{enumerate}
    \item To analyze how different quantization levels (8-bit, 4-bit, 3-bit, 2-bit) and different quantization types affect accuracy, stability, and hallucination.  
    \item To develop quantization-aware multi-pass prompt verification framework to reduce hallucinations and improve thematic alignment for expert and non-expert inputs in low precision models. 
    \item To evaluate if quantized low-precision models using this framework are capable of achieving performance levels suitable for practical use.
    
\end{enumerate}

This paper makes three key contributions. First, it provides the first systematic evaluation of multiple quantization levels (8-bit, 4-bit, 3-bit, and 2-bit) and quantization types for qualitative transcript analysis under hallucination-aware metrics. Second, it introduces a quantization-aware multi-pass verification workflow specifically designed to stabilize low-bit local LLM deployments. Third, it empirically demonstrates that structured verification can compensate for precision loss in resource-constrained settings, enabling practical qualitative analysis without increasing model size.


\section{Related Works}

Early work on post-training quantization for transformers focused on high accuracy at low bit-width. GPTQ enables effective 3–4 bit quantization with minimal accuracy loss \cite{frantar2023gptq}. It shows that careful weight update during quantization can preserve generation quality while greatly reducing memory. SmoothQuant enables 8-bit weight and activation quantization (W8A8) by migrating quantization difficulty from activations to weights \cite{xiao2023smoothquant}. SmoothQuant enables near-lossless 8-bit quantization. AWQ proposes activation-aware weight quantization for low-bit (INT3/4) LLMs \cite{lin2024awq}. AWQ enables stable 3–4 bit compression. Our work uses AWQ-style formats as one of several 3-bit configurations in the experiments. SpQR achieves near-lossless compression at very low bit widths. \cite{dettmers2023spqr}. It achieves near-lossless compression across model scales and can even reach 2-bit effective precision with small perplexity loss. Later evaluations showed that SpQR can outperform GPTQ at very low bit-widths \cite{jin2024quantization}. Our study includes SPQR for 2-bit quantization, AQW for 3-bit quantization, GPTQ for 4-bit quantization, and SmoothQuant for 8-bit quantization, but instead of focusing on perplexity or reasoning benchmarks, we test their behavior for thematic coding, frequency analysis, and hallucination with noisy, non-expert language.




Jin et al. \cite{li2023llmmq} provided a comprehensive evaluation of multiple quantization strategies across ten benchmarks, covering knowledge, alignment, and efficiency. They show that 4-bit models can stay close to full precision on many tasks, while 2-bit models often degrade strongly. They also highlighted engineering challenges, such as speed and hardware constraints, when deploying quantized LLMs. Our study is inspired by this structured evaluation but focuses on expert/non-expert coding, qualitative analysis, and hallucination-aware metrics rather than generic Question-Answer (QA) or reasoning benchmarks. Huang et al.\cite{huang2025hallucination}  surveyed hallucination in LLMs. They discussed causes such as exposure bias, miscalculated confidence, and prompt design issues, and also provided taxonomies. They outlined mitigation strategies, which include verification, retrieval, and better supervision, but do not study quantization as a factor. Our work connects these two areas by showing how low-bit quantization can affect hallucination in qualitative tasks, and by evaluating a concrete multi-pass verification scheme tailored to quantized models.

Recent work focused on how LLMs can be used in qualitative research and coding workflows. Adeseye et al. \cite{adeseye2025promptframework} evaluated LLMs for thematic analysis, frequency extraction, and impact evaluation on anonymized interview transcripts. They compared cost, throughput, hallucination rate, and accuracy for models of different sizes. Their study motivates the use of local LLMs for privacy-sensitive qualitative work. 

While multi-pass prompting and self-verification strategies have been explored in prior prompt-engineering literature, including self-refinement \cite{feng2025tear}, chain-of-thought verification \cite{xing2025incorporating}, and grounded generation approaches \cite{lertvittayakumjorn2025towards}, as well as critique-based prompting \cite{lin2024criticbench}, constrained structured generation \cite{banerjee2025crane}, and evidence-grounded extraction frameworks \cite{hossain2026biogen}, these works primarily focus on improving reasoning accuracy in full-precision models or large API-based systems. In addition, retrieval-augmented generation (RAG) methods enhance factual reliability by incorporating external knowledge sources during inference \cite{hwang2025retrieval}. However, these approaches do not examine how numerical precision constraints introduced by aggressive post-training quantization alter model reliability, hallucination tendencies, or thematic stability in qualitative analysis tasks.

In contrast, this study investigates verification prompting under extreme low-bit quantization (2–8 bit) in local deployment settings, where representational capacity is significantly reduced. Furthermore, unlike prior reasoning-centric benchmarks, our evaluation is conducted on long-form qualitative interview transcripts requiring thematic coding, frequency extraction, and clustering validation. Rather than augmenting the model with external retrieval or additional supervision, the proposed framework enforces strict transcript-grounded validation through iterative internal verification, isolating the effects of quantization from knowledge augmentation. This combination of quantization-aware evaluation, domain-specific qualitative analysis, and structured verification workflow constitutes the primary novelty of this work.

\section{Methodology}

This study follows a multi-phase methodology designed to examine how different quantization levels affect the accuracy of qualitative thematic coding and frequency analysis. It also evaluates how the proposed quantization-aware prompts and multi-pass verification process improve accuracy across low-bit models. The full workflow is illustrated in Figure \ref{fig:multiphasemethodology}. The diagram shows the complete process flow of this research, from human coding and ground-truth creation to quantized model analysis and final evaluation. The brief discussion of each phase can be seen below:

\begin{figure}
    \centering
    \includegraphics[width=0.9\linewidth]{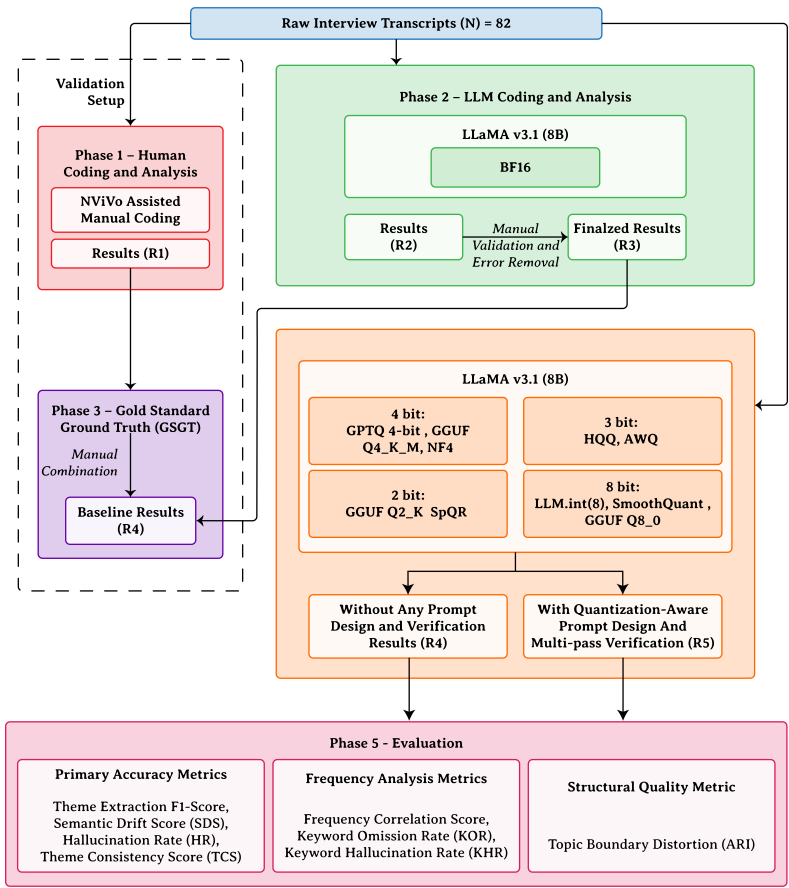}
   \vspace{-0.3CM}
    \caption{Multi-phase Methodology}
    
    \label{fig:multiphasemethodology}
    \vspace{-0.7Cm}
\end{figure}

Phase 1 – Human Coding: First, two researchers coded all transcripts in NVivo. They identified themes, keywords, quotes, and frequencies. As a result, this created set R1.

Phase 2 – BF16 LLM Coding: Next, LLaMA-3.1-8B (BF16) analyzed the same transcripts. Afterward, 2 researchers checked and corrected the results. Consequently, this created set R3.

Phase 3 – Subsequently, R1 (the consolidated output of the two researchers) was combined with R3. During this integration process, only manually verified and mutually agreed-upon findings were retained. The resulting output formed the Gold-Standard Ground Truth (GSGT), denoted as R4. The primary purpose of this step was to mitigate the risk of human omission bias. While researchers may unintentionally overlook certain relevant insights, LLM-assisted analysis (R3) may surface additional findings. However, to preserve methodological rigor and reliability, any LLM-identified insight was incorporated into the GSGT only after manual verification and agreement by both researchers. This ensured that the final baseline reflected a rigorously validated and comprehensive representation of the qualitative findings.

Phase 4 – Quantized Evaluation: Subsequently, quantized models (8-, 4-, 3-, and 2-bit) analyzed the transcripts. At this stage, tests were done without and with multi-pass verification. Accordingly, the outputs were raw results and verified results (R5).

Phase 5 – Evaluation: Finally, all outputs were compared with the GSGT. In this step, metrics measured accuracy, drift, hallucination, stability, frequency, and clustering (see Section \ref{subsec:EvaluationMetrics}). Complete information about the implementation and gold standard ground truth are provided in appendix 8.1 and 8.2.

\subsection{Quantized Models}

Lower-precision versions of the model are created using several quantization methods. The 8-bit group includes SmoothQuant, INT8, and GGUF Q8\_0. These methods quantize both the weights and the activations into 8-bit integers, which keeps accuracy high while reducing memory use. 
The 4-bit group includes Q4\_K\_M, NF4, and GPTQ-4bit. Q4\_K\_M quantizes the weights into 4-bit integers and applies per-group scaling to stabilise the activations. The activations themselves stay in higher precision, but the grouped scaling helps keep them within a safe range during inference. NF4 quantizes the weights into 4-bit floating-point values, while activations stay in higher precision. GPTQ-4bit quantizes the weights into 4-bit integers using layer-wise error correction and also keeps activations in higher precision. 
The 3-bit methods, HQQ and AWQ, quantize the weights into 3-bit integers. HQQ protects important weights using Hessian-based selection, while AWQ uses activation-aware scoring to decide which weights need more precision. Both methods keep activations in higher precision to avoid instability. 
The 2-bit methods, SPQR and GGUF Q2\_K, quantize the weights into 2-bit integers for extreme compression. SPQR also uses sparsity information to reduce quantization error. GGUF Q2\_K applies grouped 2-bit quantization for efficient local inference. Both methods keep activations at higher precision because 2-bit activation quantization is not feasible for transformers.

\subsection{Dataset}

The dataset used in this study consists of 82 semi-structured interviews. The participants came from different sectors, including NGOs, companies, universities, government agencies, and healthcare organizations. Each interview lasted between 45 and 60 minutes. The interviews produced long transcripts, ranging from 8,000 to 13,000 words. The interviews focused on privacy concerns related to introducing gamification inside organizations. Participants were asked about their level of knowledge in privacy and data protection. Those who said they were moderately familiar or very familiar were classified as experts (n = 33). The remaining participants, who had lower knowledge or no formal background in privacy, were classified as non-experts (n = 49). The dataset contains a mix of clear expert terminology and non-standard terms used by non-experts. This makes the dataset suitable for studying how quantized LLMs handle different levels of language clarity. It also allows evaluation of how models respond to both well-defined technical expressions and everyday descriptions of privacy concerns.

\subsection{Evaluation Metrics}
\label{subsec:EvaluationMetrics}

This study uses eight evaluation metrics to measure the impact of LLM quantization on qualitative thematic coding and frequency analysis. Each metric is explained below:


\subsubsection{Theme Extraction F1-Score (F1) (Higher is Better)} 
This measures how accurately the model extracts themes compared to the Gold-Standard Ground Truth (GSGT). It is measured by comparing model-generated themes with the GSGT to compute precision, recall, and F1 score. The rating scale used is 0-1, where 0 is poor and 1 is perfect. The formula used is:

{\scriptsize\[
\text{Precision} = \frac{TP}{TP + FP}, \quad
\text{Recall} = \frac{TP}{TP + FN}
\]}

{\scriptsize\[
\text{F1} = \frac{2 \times (\text{Precision} \times \text{Recall})}
{\text{Precision} + \text{Recall}}
\]}

Where $TP$ = correct themes, $FP$ = incorrect themes, $FN$ = missed themes. In implementation, theme matching was conducted using a two-stage procedure. First, exact string matching was applied after lowercasing and stop-word normalization. If no exact match was found, cosine similarity between theme embeddings was computed using a fixed embedding model. A similarity threshold of 0.80 was used to determine semantic equivalence. Themes exceeding the threshold were counted as true positives; unmatched model themes were counted as false positives, and unmatched GSGT themes were counted as false negatives. Sub-theme matching followed the same procedure.


\subsubsection{Semantic Drift Score (SDS) (Lower is Better)}
This measures how much the meaning of the model output drifts away from the true meaning of the transcript. It is measured by computing the cosine similarity between embeddings of model output and ground truth. The rating scale used is between to 0-1 where 0 is no drift (best) and 1 is severe drift (worst). The formula used is:

{\scriptsize\[
\text{SDS} = 1 - \cos(\theta)
\]}

{\scriptsize\[
\cos(\theta) =
\frac{E_{\text{LLM}} \cdot E_{\text{GSGT}}}
{\|E_{\text{LLM}}\|\|E_{\text{GSGT}}\|}
\]}

Embeddings for both model outputs and GSGT references were generated using the same fixed embedding model to ensure consistency across comparisons.

\subsubsection{Hallucination Rate (HR) (Lower is Better) } This measures how much incorrect or unsupported content the model adds. All generated statements are checked against the transcript. Unsupported statements are counted as hallucinations. The rating scale used is between to 0-1 where 0 is no hallucinations (best) and 1 is all content is hallucinated. The formula used is:

{\scriptsize\[
HR = \frac{\text{Number of hallucinated statements}}
{\text{Total generated statements}}
\]}
A generated statement was classified as hallucinated if it introduced information not explicitly supported by the transcript. Validation was performed using a hybrid procedure: (1) direct string containment check, (2) embedding-based semantic similarity between generated quotes and transcript segments (threshold = 0.80), and (3) manual verification by human coders when automated matching was inconclusive. Statements failing validation were counted as hallucinations.

\subsubsection{Theme Consistency Score (TCS) (Higher is Better)} 
This measures the stability of the model when producing themes across repeated runs. The same prompt is run multiple times. Cosine similarity between outputs is averaged. The rating scale used is between to 0-1 where 0 is inconsistent and 1 is highly stable. The formula used is:

\vspace{-0.1CM}
{\scriptsize

\[
TCS = \frac{1}{N}\sum_{i=1}^{N} \cos(\theta_i)
\]
}
Where $\theta_i$ is the similarity between two outputs.


\subsubsection{Frequency Correlation Score (Higher is Better)}
This measures how well the model preserves the true frequency pattern of keywords. It is measured by computing the correlation between model keyword counts and ground-truth counts. The rating scale used is 0-1, where 0 is no correlation and 1 is perfect correlation. The formula used is:

{\scriptsize\[
r =
\frac{\sum (x_i - \bar{x})(y_i - \bar{y})}
{\sqrt{\sum (x_i - \bar{x})^2}
 \sqrt{\sum (y_i - \bar{y})^2}}
\]}

Where $x_i$ and $y_i$ are keyword frequencies. Theme and sub-theme counts were extracted directly from structured JSON outputs. Counts were normalized before correlation computation to ensure comparability across transcripts.


\subsubsection{Keyword Omission Rate (KOR) (Lower is Better)}
This measures the number of important keywords the model fails to extract. It is measured by counting all ground-truth keywords missing from the model's output. The rating scale used is between to 0-1 where 0 is no omissions (best) and 1 is all keywords omitted (worst). The formula used is:

{\scriptsize\[
KOR = \frac{\text{Missed Ground Truth Keywords}}
{\text{Total Ground Truth Keywords}}
\]}

In implementation, the Ground-Standard Ground Truth (GSGT) keyword list was first normalized through lowercasing, lemmatization, and removal of stop words to ensure consistency. Model-generated keywords were processed using the same normalization pipeline. A keyword was considered correctly extracted if it matched the GSGT keyword either through exact string matching or through semantic similarity using cosine similarity between embeddings (threshold = 0.80). Ground-truth keywords that did not meet either matching criterion were counted as omissions. The omission rate was computed at the transcript level and then averaged across all transcripts.


\subsubsection{Keyword Hallucination Rate (KHR) (Lower is Better)}
This measures the number of extra or invented keywords added by the model. It is measured by counting all keywords generated by the model that do not appear in the transcript. The rating scale used is between to 0-1 where 0 is no hallucinated keywords (best) and 1 is all keywords hallucinated (worst). The formula used is:

{\scriptsize\[
KHR = \frac{\text{Invented Keywords}}
{\text{Total Model Keywords}}
\]}

A keyword was classified as hallucinated if it appeared in the model output but could not be matched to any transcript segment or GSGT keyword. Validation followed a two-step process: (1) direct lexical containment within the transcript text after normalization, and (2) semantic similarity comparison using cosine similarity (threshold = 0.80) between generated keywords and transcript embeddings. Keywords failing both checks were counted as hallucinated. The hallucination rate was computed per transcript and averaged across the dataset.


\subsubsection{Adjusted Rand Index (ARI) (Higher is Better)}
This measures how accurately the model groups segments into the correct thematic clusters. It is computed by comparing the model's theme clustering structure with the ground truth using the Adjusted Rand Index. The rating scale used is between 0-1, where 1 means a perfect match, and 1 means random clustering. However, it is also possible to get $< 0$ when the results is worse than random. The formula used is:

\vspace{-0.1CM}
{\scriptsize\[
ARI = \frac{RI - E[RI]}{\max(RI) - E[RI]}
\]}

\vspace{-0.1CM}
Where $RI$ is the Rand Index computed from pairs of matching and non-matching items.


\section{Multi-Pass Prompt Verification}

\begin{figure}
    \centering
    \includegraphics[width=0.8\linewidth]{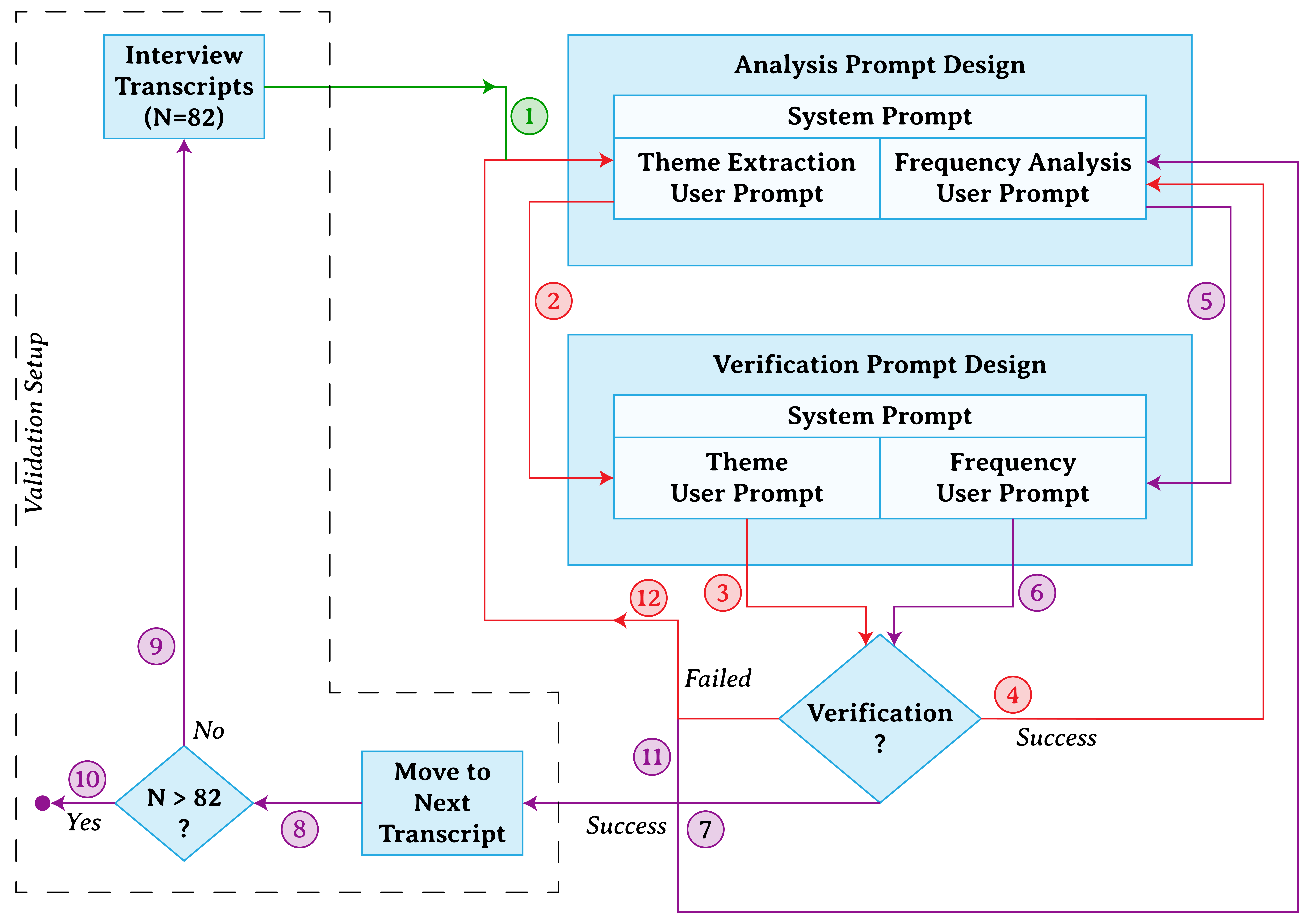}
    \vspace{-0.5cm}
    \caption{Multi-Pass Prompt Verification Framework (Note: The numbers shown in the arrows indicates order of functional flow)}
    \vspace{-0.4cm}
    \label{fig:multipass}
    \vspace{-0.4CM}
\end{figure}

Quantized models reduce numerical precision to lower memory requirements and speed up inference. They use fewer bits to represent numerical values. This improves computational efficiency but can affect model accuracy. Reduced precision may introduce numerical errors that lead to unstable or inconsistent outputs for identical inputs. These models often struggle with long or complex instructions. They can have difficulty processing abstract reasoning tasks. As numerical precision decreases, these limitations become more pronounced. Overall, lower precision can weaken reasoning performance and reduce the reliability of generated results.
Figure \ref{fig:multipass} illustrates the multi-pass prompt verification framework. In this process, interview transcripts are processed step by step.

First, the system extracts themes and sub-themes. Next, these themes are verified using a second prompt. if errors are found, the process is repeated so that only confirmed themes are accepted. After verification of the themes, the system counts the frequency of each theme and sub-theme before moving to the verification. Again, verification of the frequency count should be confirmed before moving to the next transcript and then repeating the process. Overall, this framework ensures higher accuracy. Additionally, it reduces inconsistent outputs, improving the stability and reliability of low-bit models. Figure \ref{fig:multipass} has two prompt design types: Analysis and Verification. Both types have a structured system prompt used for both theme extraction and frequency analysis \cite{adeseye2025promptframework}. 
The system prompt structure for analysis (theme extraction and frequency analysis) prompt design is given below:

\vspace{0.5em}
{\scriptsize
\hspace{1.5em}
\begin{minipage}{\dimexpr\linewidth-1.5em}
\begin{verbatim}
You are a careful qualitative analysis model.
Do not guess. Use only the transcript.
Keep sentences short. Follow the steps.
Tasks:
Identify themes and sub-themes.
Give short explanations.
Provide exact supporting quotes.
Do not add outside knowledge.
\end{verbatim}
\end{minipage}
}

\vspace{0.5em}

The user prompts structure utilized for themes and sub-themes extraction is given below:

\vspace{0.5em}
{\scriptsize
\hspace{1.5em}
\begin{minipage}{\dimexpr\linewidth-1.5em}
\begin{verbatim}
Analyze the transcript using the rules.
Return JSON: {
"themes": [{
"theme_id": "T1",
"description": "...",
"subthemes": [{
"subtheme_id": "ST1",
"description": "...",
"quotes": ["...", "..."] } ] } ] }
\end{verbatim}
\end{minipage}
}
\vspace{0.5em}

The user prompts structure used to count the frequency are given below:

\vspace{0.5em}
{\scriptsize
\hspace{1.5em}
\begin{minipage}{\dimexpr\linewidth-1.5em}
\begin{verbatim}
Use the verified themes and sub-themes.
Count how many times each appears.
Return JSON: {
"theme_frequencies": [ {
"theme_id": "T1",
"count": 0,
"subthemes": [ {
"subtheme_id": "ST1",
"count": 0 } ] } ] }
\end{verbatim}
\end{minipage}
}
\vspace{0.5em}

Overall, this analysis prompt design structure reduces cognitive load on low-bit models. hereby, limiting hallucination and improving stability, which ensures accurate, repeatable theme extraction and frequency count.
The system prompt structure used for the verification of the theme extraction and frequency analysis is provided below:

\vspace{0.5em}
{\scriptsize
\hspace{1.5em}
\begin{minipage}{\dimexpr\linewidth-1.5em}
\begin{verbatim}
You are a careful verification model.
Do not guess. Use only the transcript and JSON.
Keep sentences short. Follow the steps.
Tasks:
Check all themes and sub-themes.
Remove unsupported items.
Check quotes and counts.
Do not add new content.
\end{verbatim}
\end{minipage}
}
\vspace{0.5em}

The user prompts structure used to verify the extracted themes and sub-themes is given below:

\vspace{0.5em}
{\scriptsize
\hspace{1.5em}
\begin{minipage}{\dimexpr\linewidth-1.5em}
\begin{verbatim}
Verify themes and sub-themes using the rules.
Keep only items supported by the transcript.
Remove hallucinated themes, sub-themes, and quotes.
Return updated JSON.
\end{verbatim}
\end{minipage}
}
\vspace{0.5em}

The user prompts structure used to verify the frequency count is given below:

\vspace{0.5em}
{\scriptsize
\hspace{1.5em}
\begin{minipage}{\dimexpr\linewidth-1.5em}
\begin{verbatim}
Verify all frequency counts.
Use only the transcript.
Remove unsupported counts.
Update the "count" fields only.
Do not add new themes or sub-themes.
Return updated JSON.
\end{verbatim}
\end{minipage}
}
\vspace{0.25em}

Overall, verification reduces errors. As a result, it lowers hallucination risk and improves the reliability of low-bit models.

\section{Validation}

The validation compares each quantized model against the Gold-Standard Ground Truth using three groups of metrics (subsection \ref{subsec:EvaluationMetrics}). Both expert (E) and non-expert (N) inputs are included to observe the differences caused by terminology clarity. Each quantized model was tested in two ways: without the proposed method (Before) and with the quantization-aware prompt design and multi-pass verification (After). Table \ref{tab:quant_p_vs_n} presents the full validation results for all quantization levels and quantization types.

\begin{table*}[htbp]
\centering
\scriptsize   
\caption{Performance of LLaMA 3.1 (8B) Under Quantization for Expert (E) and Non-Expert (N) Prompting, Before and After Applying Quantization-Aware Multi-Pass Prompt Verification}
\vspace{-0.3cm}
\label{tab:quant_p_vs_n}
{\scriptsize
\resizebox{\linewidth}{!}{
\begin{tabular}{lcccccccccccccccc}
\toprule
\textbf{Model / Condition} 
& \textbf{F1$_E$} & \textcolor{blue}{\textbf{F1$_N$}}
& \textbf{SDS$_E$} & \textcolor{blue}{\textbf{SDS$_N$}}
& \textbf{HR$_E$} & \textcolor{blue}{\textbf{HR$_N$}}
& \textbf{TCS$_E$} & \textcolor{blue}{\textbf{TCS$_N$}}
& \textbf{Freq$_E$} & \textcolor{blue}{\textbf{Freq$_N$}}
& \textbf{KOR$_E$} & \textcolor{blue}{\textbf{KOR$_N$}}
& \textbf{KHR$_E$} & \textcolor{blue}{\textbf{KHR$_N$}}
& \textbf{ARI$_E$} & \textcolor{blue}{\textbf{ARI$_N$}} \\
\midrule

\textbf{Human (Reference)} 
& 0.92 & \textcolor{blue}{0.80}
& 0.02 & \textcolor{blue}{0.30}
& 0.00 & \textcolor{blue}{0.00}
& 0.95 & \textcolor{blue}{0.89}
& 0.94 & \textcolor{blue}{0.62}
& 0.03 & \textcolor{blue}{0.10}
& 0.00 & \textcolor{blue}{0.00}
& 0.90 & \textcolor{blue}{0.71} \\
\midrule

\textbf{BF16 (Before)} 
& 0.90 & \textcolor{blue}{0.65}
& 0.04 & \textcolor{blue}{0.24}
& 0.02 & \textcolor{blue}{0.22}
& 0.93 & \textcolor{blue}{0.68}
& 0.93 & \textcolor{blue}{0.68}
& 0.04 & \textcolor{blue}{0.24}
& 0.02 & \textcolor{blue}{0.22}
& 0.88 & \textcolor{blue}{0.63} \\

\textbf{BF16 (After)}  
& 0.92 & \textcolor{blue}{0.82}
& 0.03 & \textcolor{blue}{0.11}
& 0.01 & \textcolor{blue}{0.09}
& 0.95 & \textcolor{blue}{0.85}
& 0.95 & \textcolor{blue}{0.85}
& 0.03 & \textcolor{blue}{0.11}
& 0.01 & \textcolor{blue}{0.09}
& 0.90 & \textcolor{blue}{0.80} \\
\midrule

\multicolumn{17}{c}{\textbf{8-bit Quantization}} \\
\midrule

\textbf{SmoothQuant (Before)} 
& 0.88 & \textcolor{blue}{0.63}
& 0.06 & \textcolor{blue}{0.26}
& 0.04 & \textcolor{blue}{0.24}
& 0.91 & \textcolor{blue}{0.66}
& 0.92 & \textcolor{blue}{0.66}
& 0.06 & \textcolor{blue}{0.26}
& 0.04 & \textcolor{blue}{0.24}
& 0.86 & \textcolor{blue}{0.61} \\

\textbf{SmoothQuant (After)}  
& 0.91 & \textcolor{blue}{0.81}
& 0.04 & \textcolor{blue}{0.12}
& 0.02 & \textcolor{blue}{0.10}
& 0.94 & \textcolor{blue}{0.84}
& 0.94 & \textcolor{blue}{0.84}
& 0.04 & \textcolor{blue}{0.12}
& 0.02 & \textcolor{blue}{0.10}
& 0.89 & \textcolor{blue}{0.79} \\
\midrule

\textbf{INT8 (Before)} 
& 0.84 & \textcolor{blue}{0.59}
& 0.10 & \textcolor{blue}{0.30}
& 0.07 & \textcolor{blue}{0.27}
& 0.85 & \textcolor{blue}{0.60}
& 0.87 & \textcolor{blue}{0.62}
& 0.10 & \textcolor{blue}{0.30}
& 0.07 & \textcolor{blue}{0.27}
& 0.80 & \textcolor{blue}{0.55} \\

\textbf{INT8 (After)}  
& 0.87 & \textcolor{blue}{0.77}
& 0.08 & \textcolor{blue}{0.16}
& 0.04 & \textcolor{blue}{0.12}
& 0.88 & \textcolor{blue}{0.78}
& 0.89 & \textcolor{blue}{0.79}
& 0.07 & \textcolor{blue}{0.15}
& 0.04 & \textcolor{blue}{0.12}
& 0.83 & \textcolor{blue}{0.73} \\
\midrule

\textbf{GGUF Q8\_0 (Before)} 
& 0.83 & \textcolor{blue}{0.58}
& 0.11 & \textcolor{blue}{0.31}
& 0.08 & \textcolor{blue}{0.28}
& 0.84 & \textcolor{blue}{0.59}
& 0.86 & \textcolor{blue}{0.61}
& 0.11 & \textcolor{blue}{0.31}
& 0.08 & \textcolor{blue}{0.28}
& 0.79 & \textcolor{blue}{0.54} \\

\textbf{GGUF Q8\_0 (After)}  
& 0.86 & \textcolor{blue}{0.76}
& 0.09 & \textcolor{blue}{0.17}
& 0.05 & \textcolor{blue}{0.13}
& 0.87 & \textcolor{blue}{0.77}
& 0.88 & \textcolor{blue}{0.78}
& 0.08 & \textcolor{blue}{0.16}
& 0.05 & \textcolor{blue}{0.13}
& 0.82 & \textcolor{blue}{0.72} \\
\midrule

\multicolumn{17}{c}{\textbf{4-bit Quantization}} \\
\midrule

\textbf{Q4\_K\_M (Before)} 
& 0.83 & \textcolor{blue}{0.58}
& 0.12 & \textcolor{blue}{0.32}
& 0.07 & \textcolor{blue}{0.27}
& 0.85 & \textcolor{blue}{0.60}
& 0.86 & \textcolor{blue}{0.61}
& 0.10 & \textcolor{blue}{0.30}
& 0.07 & \textcolor{blue}{0.27}
& 0.78 & \textcolor{blue}{0.53} \\

\textbf{Q4\_K\_M (After)}  
& 0.87 & \textcolor{blue}{0.77}
& 0.08 & \textcolor{blue}{0.16}
& 0.04 & \textcolor{blue}{0.12}
& 0.89 & \textcolor{blue}{0.79}
& 0.89 & \textcolor{blue}{0.79}
& 0.07 & \textcolor{blue}{0.15}
& 0.04 & \textcolor{blue}{0.12}
& 0.83 & \textcolor{blue}{0.73} \\
\midrule

\textbf{NF4 (Before)} 
& 0.86 & \textcolor{blue}{0.61}
& 0.09 & \textcolor{blue}{0.29}
& 0.05 & \textcolor{blue}{0.25}
& 0.88 & \textcolor{blue}{0.63}
& 0.89 & \textcolor{blue}{0.64}
& 0.08 & \textcolor{blue}{0.28}
& 0.05 & \textcolor{blue}{0.25}
& 0.82 & \textcolor{blue}{0.57} \\

\textbf{NF4 (After)}	 
& 0.89 & \textcolor{blue}{0.79}
& 0.06 & \textcolor{blue}{0.14}
& 0.03 & \textcolor{blue}{0.11}
& 0.91 & \textcolor{blue}{0.81}
& 0.91 & \textcolor{blue}{0.81}
& 0.06 & \textcolor{blue}{0.14}
& 0.03 & \textcolor{blue}{0.11}
& 0.85 & \textcolor{blue}{0.75} \\
\midrule

\textbf{GPTQ-4bit (Before)} 
& 0.80 & \textcolor{blue}{0.55}
& 0.16 & \textcolor{blue}{0.36}
& 0.09 & \textcolor{blue}{0.29}
& 0.82 & \textcolor{blue}{0.57}
& 0.83 & \textcolor{blue}{0.58}
& 0.13 & \textcolor{blue}{0.33}
& 0.09 & \textcolor{blue}{0.29}
& 0.76 & \textcolor{blue}{0.51} \\

\textbf{GPTQ-4bit (After)}  
& 0.84 & \textcolor{blue}{0.74}
& 0.12 & \textcolor{blue}{0.20}
& 0.07 & \textcolor{blue}{0.15}
& 0.86 & \textcolor{blue}{0.76}
& 0.87 & \textcolor{blue}{0.77}
& 0.10 & \textcolor{blue}{0.18}
& 0.07 & \textcolor{blue}{0.15}
& 0.80 & \textcolor{blue}{0.70} \\
\midrule

\multicolumn{17}{c}{\textbf{3-bit Quantization}} \\
\midrule

\textbf{HQQ (Before)} 
& 0.62 & \textcolor{blue}{0.37}
& 0.35 & \textcolor{blue}{0.55}
& 0.20 & \textcolor{blue}{0.40}
& 0.60 & \textcolor{blue}{0.35}
& 0.68 & \textcolor{blue}{0.43}
& 0.25 & \textcolor{blue}{0.45}
& 0.20 & \textcolor{blue}{0.40}
& 0.55 & \textcolor{blue}{0.30} \\

\textbf{HQQ (After)}  
& 0.68 & \textcolor{blue}{0.58}
& 0.28 & \textcolor{blue}{0.36}
& 0.14 & \textcolor{blue}{0.22}
& 0.67 & \textcolor{blue}{0.57}
& 0.73 & \textcolor{blue}{0.63}
& 0.20 & \textcolor{blue}{0.28}
& 0.14 & \textcolor{blue}{0.22}
& 0.62 & \textcolor{blue}{0.52} \\
\midrule

\textbf{AWQ (Before)} 
& 0.58 & \textcolor{blue}{0.33}
& 0.38 & \textcolor{blue}{0.58}
& 0.23 & \textcolor{blue}{0.43}
& 0.56 & \textcolor{blue}{0.31}
& 0.64 & \textcolor{blue}{0.39}
& 0.28 & \textcolor{blue}{0.48}
& 0.23 & \textcolor{blue}{0.43}
& 0.50 & \textcolor{blue}{0.26} \\

\textbf{AWQ (After)}  
& 0.63 & \textcolor{blue}{0.53}
& 0.32 & \textcolor{blue}{0.40}
& 0.18 & \textcolor{blue}{0.26}
& 0.62 & \textcolor{blue}{0.52}
& 0.69 & \textcolor{blue}{0.58}
& 0.24 & \textcolor{blue}{0.32}
& 0.18 & \textcolor{blue}{0.26}
& 0.57 & \textcolor{blue}{0.47} \\
\midrule

\multicolumn{17}{c}{\textbf{2-bit Quantization}} \\
\midrule

\textbf{SPQR (Before)} 
& 0.50 & \textcolor{blue}{0.25}
& 0.48 & \textcolor{blue}{0.68}
& 0.30 & \textcolor{blue}{0.50}
& 0.45 & \textcolor{blue}{0.20}
& 0.55 & \textcolor{blue}{0.30}
& 0.35 & \textcolor{blue}{0.55}
& 0.30 & \textcolor{blue}{0.50}
& 0.40 & \textcolor{blue}{0.15} \\

\textbf{SPQR (After)}  
& 0.55 & \textcolor{blue}{0.45}
& 0.40 & \textcolor{blue}{0.48}
& 0.22 & \textcolor{blue}{0.30}
& 0.52 & \textcolor{blue}{0.42}
& 0.63 & \textcolor{blue}{0.53}
& 0.28 & \textcolor{blue}{0.36}
& 0.22 & \textcolor{blue}{0.30}
& 0.48 & \textcolor{blue}{0.38} \\
\midrule

\textbf{GGUF Q2\_K (Before)} 
& 0.46 & \textcolor{blue}{0.22}
& 0.52 & \textcolor{blue}{0.72}
& 0.34 & \textcolor{blue}{0.54}
& 0.40 & \textcolor{blue}{0.17}
& 0.50 & \textcolor{blue}{0.27}
& 0.38 & \textcolor{blue}{0.58}
& 0.34 & \textcolor{blue}{0.54}
& 0.36 & \textcolor{blue}{0.12} \\

\textbf{GGUF Q2\_K (After)}  
& 0.51 & \textcolor{blue}{0.40}
& 0.45 & \textcolor{blue}{0.52}
& 0.26 & \textcolor{blue}{0.34}
& 0.47 & \textcolor{blue}{0.38}
& 0.58 & \textcolor{blue}{0.48}
& 0.32 & \textcolor{blue}{0.40}
& 0.26 & \textcolor{blue}{0.34}
& 0.43 & \textcolor{blue}{0.34} \\

\bottomrule

\end{tabular}
} 
}
\vspace{-0.7cm}
\end{table*}

The results in Table \ref{tab:quant_p_vs_n} show clear patterns across quantization precision and user groups. Expert users (E) provide precise terminology, which helps all models perform better. Non-experts (N) use unclear language, leading to lower baseline accuracy, especially at reduced numerical precision. Without verification, performance drops sharply for 4-bit, 3-bit, and 2-bit models across all metrics. 
The proposed quantization-aware multi-pass framework significantly improves efficiency and accuracy across all tested models. For high-precision models (BF16 and 8-bit), improvements range between 10–25\% depending on the metric, pushing results close to human reference levels. For 4-bit models, the relative gains are even stronger, with improvements of 40–80\%, effectively doubling performance on several tests, especially for Non-Expert prompts. This gain makes 4-bit models competitive with 8-bit baselines while requiring far less memory and faster inference, demonstrating strong efficiency benefits. Improvements remain substantial for 3-bit and 2-bit models, although absolute accuracy remains lower. Here, the framework still reduces hallucination and stabilizes output by validating themes, matching quotes, and correcting frequency counts step by step. These checks are most valuable for unclear user inputs, where unsupported themes and incorrect statistics are removed. Overall, the framework delivers large accuracy gains without increasing model size or compute cost, making 4-bit quantized models both accurate and practically usable for qualitative analysis. This confirms that verification efficiency, rather than increased model precision alone, is key to enabling reliable low-resource LLM deployments.

\subsection{Component-Level Analysis of the Multi-Pass Framework (Ablation Study)}

To better understand which elements of the proposed framework contribute most to performance improvements, this subsection analyzes the system at the component level. The framework consists of structured prompt design, JSON output constraints, theme-level verification, quote validation, frequency verification, and iterative multi-pass execution. Although the ``Before'' and ``After'' conditions in Table 1 represent the system without and with full verification respectively, the observed metric changes allow us to infer the relative contribution of each component.

\subsubsection{Structured Prompt Design and JSON Constraint}

Structured system prompting and enforced JSON output formatting provide baseline stability even before verification is applied. In the ``Before'' condition, Theme Consistency Scores (TCS) remain relatively stable for higher precision models, such as SmoothQuant (TCS$_E$ = 0.91) and INT8 (TCS$_E$ = 0.85), indicating that structured output reduces variability across repeated runs. However, improvements in extraction accuracy and hallucination control remain limited at this stage. For example, GPTQ-4bit achieves F1$_N$ = 0.55, HQQ (3-bit) achieves F1$_N$ = 0.37, and SPQR (2-bit) achieves F1$_N$ = 0.25 prior to verification. These results indicate that structured prompting alone improves format consistency but does not sufficiently mitigate precision-induced semantic errors under aggressive quantization.

\subsubsection{Theme Verification as the Primary Driver of Improvement}

The largest performance gains are observed following theme-level verification, suggesting that this component is the dominant driver of improvement. For HQQ (3-bit), F1$_N$ increases from 0.37 to 0.58 (+21 percentage points), while Hallucination Rate (HR$_N$) decreases from 0.40 to 0.22, representing a 45\% reduction. Similarly, SPQR (2-bit) shows F1$_N$ improvement from 0.25 to 0.45 (+20 points), with HR$_N$ decreasing from 0.50 to 0.30 (40\% reduction). These substantial changes indicate that removing unsupported or weakly grounded themes during verification directly addresses hallucination and precision drift. The magnitude of improvement is most pronounced in lower-bit models, confirming that theme verification compensates for representational degradation caused by extreme compression.

\subsubsection{Quote Validation and Semantic Grounding}

Quote validation strengthens semantic alignment by ensuring that retained themes are explicitly supported by transcript content. This component contributes directly to reductions in Semantic Drift Score (SDS). For example, GPTQ-4bit SDS$_N$ decreases from 0.36 to 0.20, and NF4 SDS$_N$ decreases from 0.29 to 0.14 after verification. These reductions indicate improved semantic grounding and decreased abstraction drift. By enforcing transcript-aligned support for each extracted theme, quote validation reduces unsupported generalizations and constrains model outputs within the original discourse context.

\subsubsection{Frequency Verification and Quantitative Reliability}

Frequency verification primarily affects numerical consistency and clustering alignment. For Q4 K M (4-bit), Freq$_N$ increases from 0.61 to 0.79, while ARI$_N$ improves from 0.53 to 0.73. For GGUF Q2 K (2-bit), Freq$_N$ increases from 0.27 to 0.48, and ARI$_N$ improves from 0.12 to 0.34. These gains demonstrate that miscounted theme occurrences—common in low-precision models—are substantially corrected during the frequency verification stage. This component therefore contributes specifically to quantitative reliability rather than thematic discovery itself.

\subsubsection{Iterative Multi-Pass Stabilization}

The iterative nature of the framework further enhances stability by ensuring that only verified themes and corrected counts propagate forward. Improvements in Theme Consistency Score reflect this stabilizing effect. For example, AWQ (3-bit) TCS$_N$ increases from 0.31 to 0.52, and SPQR (2-bit) TCS$_N$ increases from 0.20 to 0.42 following verification. These improvements indicate that iterative validation reduces output variance across repeated runs, thereby increasing reproducibility under low-bit constraints.

\subsubsection{Overall Findings}

Across all quantization levels, improvements are modest for BF16 and 8-bit models (typically 5--15\% F1 gain), but substantial for 4-bit and lower precision models, where F1 improvements range between 20--40 percentage points for non-expert inputs. The largest relative gains occur in hallucination reduction (up to 45\% decrease), frequency correlation (up to 21-point increase), and ARI clustering consistency (up to 22-point increase). These findings indicate that structured prompting provides baseline stability, theme verification is the primary driver of hallucination mitigation, quote validation improves semantic grounding, frequency verification enhances quantitative alignment, and iteration strengthens reproducibility. Collectively, the results demonstrate that the framework addresses distinct error classes introduced by aggressive quantization rather than producing uniform, undifferentiated improvements.

\section{Discussion}

The study observed clear differences between expert and non-expert inputs. In particular, non-expert language proved significantly harder to analyze, even for human coders. Such inputs frequently contained informal expressions, ambiguous phrases, and loosely structured arguments. Consequently, meanings were often indirect and poorly organized, increasing interpretation time and coder disagreement. This ambiguity also contributed to higher error rates and increased hallucination tendencies in model outputs.

To ensure the reliability of the human-coded baseline, inter-annotator agreement (IAA) was calculated between two independent NVivo coders prior to constructing the Gold-Standard Ground Truth (GSGT). Cohen’s Kappa ($\kappa$) for theme identification was 0.87 (87\% agreement), and for keyword/frequency coding $\kappa = 0.83$ (83\% agreement), indicating strong to near-perfect agreement according to established interpretation thresholds. The overall percentage agreement across all coded segments was 89.4\%. These results confirm that the ground truth used for evaluation is statistically reliable and robust.

In contrast, Large Language Models (LLMs) demonstrated improved performance when processing expert inputs. Expert-authored text typically employed standardized terminology, well-defined concepts, and structured reasoning. This linguistic clarity aligns closely with the distribution and structure of model training data, thereby facilitating more accurate theme extraction and frequency estimation.

Notably, the performance gap widened as numerical precision decreased. FP16 and 8-bit quantized models maintained sufficient representational capacity to handle ambiguous or loosely structured text. However, 3-bit and 2-bit models exhibited substantial degradation in reliability, with frequent failures in thematic consistency and frequency accuracy.

The proposed multi-pass prompt verification framework improved performance across all quantization levels. Following verification, F1 scores increased consistently, while Semantic Drift Score (SDS) and hallucination rates decreased. Additionally, improvements were observed in theme consistency and frequency accuracy metrics. The magnitude of improvement varied across quantization levels. For BF16 and 8-bit models, gains were modest due to their already stable baseline performance. In contrast, substantial improvements were recorded for 4-bit and lower-bit models.

Without the verification framework, many outputs from low-bit models were practically unusable for analytical purposes. After applying multi-pass verification, these outputs achieved sufficient reliability for structured qualitative analysis. The effect was particularly pronounced for non-expert data, where verification effectively removed unsupported themes, clarified ambiguous interpretations, and corrected frequency counting inconsistencies. This directly mitigated hallucination triggered by vague or informal language.

Given that the GSGT was constructed with high inter-annotator reliability ($\kappa > 0.80$ and agreement $> 85\%$), the observed improvements in F1 score, SDS, hallucination rate, and clustering consistency can be attributed to genuine model enhancement rather than annotation variability. This strengthens the internal validity of the comparative quantization evaluation.

Although quantization methods continue to vary in intrinsic quality, the multi-pass verification framework consistently improves stability and reliability across all configurations. Importantly, weaker quantization methods demonstrated enhanced robustness after verification. Overall, the framework is essential for ensuring analytical usability under computational constraints. It enables low-bit models to remain operationally viable while protecting accuracy, particularly in the analysis of non-expert qualitative text.

It is important to note that, although iterative verification prompting shares conceptual similarities with existing self-refinement and grounding strategies, its application under aggressive low-bit quantization introduces unique stability challenges not addressed in prior literature. The empirical findings demonstrate that verification efficiency becomes increasingly critical as numerical precision decreases, revealing a precision–verification trade-off that has not been systematically explored in previous prompt-engineering studies.

\section{Conclusion}

This study proposed a quantization-aware multi-pass prompt verification framework to improve qualitative analysis using quantized LLaMA-3.1 (8B). The framework applies structured prompts and repeated verification steps to reduce errors and hallucinations. It ensures that model accuracy remains close to BF16 and human reference results, especially for 8-bit and 4-bit models. Without verification, low-bit models produced many unstable outputs and frequent errors. With the framework, extracted themes are cross-checked, supporting quotes are validated, and frequency counts are corrected step by step. This process clearly improves output stability and consistency at all precision levels.
The proposed framework makes the 4-bit model competitive and also benefits the 3-bit and 2-bit models by reducing baseline errors, although their overall performance remains limited. The framework is especially important for non-expert text, which often uses unclear or inconsistent language leading to poor performance in all models. In this case, multi-pass verification removes unsupported themes and corrects miscounts, thereby reducing hallucinations caused by vague inputs. Together, these results show that the proposed framework enables faster, more accurate, and privacy-preserving LLM systems for reliable qualitative research using resource-efficient quantized models.
Despite these strengths, some limitations remain. This study tested only the LLaMA-3.1 (8B) model, selected because earlier research \cite{adeseye2025efficientprompt} showed it to be one of the most competitive local models with no major loss in accuracy compared to larger models. The dataset also came from a single qualitative domain, which may limit generalization. Future work should test additional models such as Gemma, Mistral, and Phi, and explore larger datasets as well as mixed-precision and hybrid quantization methods.

\section{Declaration on the Use of Generative AI}
Language editing and grammar-checking tools were used to improve clarity and readability of the manuscript.






%

\section*{Appendix}

\subsection{Implementation and Reproducibility Details}

To ensure clarity and full reproducibility, this subsection details the complete inference pipeline, evaluation protocol, computational setup, and ground-truth construction methodology.

\subsubsection{Inference Stack and Tooling}

All experiments were conducted using locally hosted LLaMA-3.1 (8B) models deployed via \texttt{llama.cpp}. Quantized models were loaded using their respective formats (SmoothQuant, GPTQ, AWQ, HQQ, SPQR, and GGUF variants). No external APIs were used. The same hardware configuration was maintained across all quantization levels to ensure experimental fairness.

Experiments were executed on an NVIDIA RTX 4090 (24GB VRAM), 64GB DDR5 RAM, AMD Ryzen 9 7950X CPU, running Ubuntu 22.04 LTS. The \texttt{llama.cpp} commit version \texttt{b6d6c5f} (August 2025 stable branch) was used consistently across all runs, compiled with CUDA acceleration enabled. CUDA 12.2 and GCC 11.4.0 were used, and compiler settings remained identical for all experiments.

Average inference time per transcript (single-pass) was 41.8 seconds (8-bit), 38.6 seconds (4-bit), 36.9 seconds (3-bit), and 35.4 seconds (2-bit). Average verification cycle time was 17.2 seconds per pass. 

\subsubsection{Decoding Parameters}

To ensure deterministic and stable evaluation, decoding parameters were fixed across all experiments. Temperature was set to 0.2 to reduce randomness while preserving minimal generation flexibility. Top-p was set to 0.9, and top-k was fixed at 40. Maximum generation length was set to 2048 tokens to prevent truncation of thematic outputs. These parameters were held constant for all quantization configurations.

\subsubsection{Number of Runs and Stability Measurement}

For metrics requiring stability assessment (Theme Consistency Score), each prompt configuration was executed five independent times using identical decoding parameters. Cosine similarity was computed pairwise across runs and averaged to obtain the final TCS value. All reported stability metrics represent the mean over these repeated executions.
To evaluate statistical significance between quantization levels and verification strategies, paired Wilcoxon signed-rank tests were conducted. Effect sizes (Cohen’s d) are additionally reported. Improvements were considered statistically significant at $p < 0.05$.

\subsubsection{Embedding Model for Semantic Metrics}

All embedding-based computations, including Semantic Drift Score (SDS) and semantic theme matching, were performed using a fixed embedding model to ensure consistency. Embeddings were normalized prior to cosine similarity computation.
Specifically, the embedding model used was \texttt{BAAI/bge-large-en-v1.5} (1024-dimensional embeddings), and cosine similarity threshold for semantic alignment was fixed at 0.80. A sensitivity analysis was additionally performed at thresholds 0.70 and 0.90 to assess robustness of semantic matching criteria. Results indicate that relative performance trends remained stable across threshold variations.

\subsubsection{Transcript Handling Strategy}

Interview transcripts ranged from 8,000 to 13,000 words. To ensure compatibility with model context limits, transcripts were processed in segmented windows while preserving logical continuity.
Segmentation was implemented using a sliding window of 4096 tokens with an overlap of 512 tokens to prevent boundary-induced theme loss. Each segment was analyzed independently, and theme aggregation followed a union-merge strategy where semantically equivalent themes across segments were consolidated using embedding similarity matching.

No summarization or preprocessing that altered semantic content was applied beyond normalization (lowercasing and removal of non-semantic tokens).

\subsubsection{Quantization Configuration Details}

Each quantization level followed the configuration standards defined by its respective implementation. For weight-only quantization methods (GPTQ, AWQ, HQQ, SPQR), activations were retained in FP16 precision as specified by their original frameworks. For SmoothQuant and INT8, both weights and activations were quantized according to published configurations. Default hyperparameters recommended by original implementations were used to ensure comparability with prior literature.

\subsubsection{Verification Stopping Criteria}

The multi-pass verification framework operated in two stages: theme validation and frequency validation. A verification pass was repeated only if unsupported themes or inconsistent frequency counts were detected. The process terminated when no additional unsupported elements were identified or when two consecutive passes produced identical outputs.
Across all transcripts, convergence was achieved within two verification cycles in 93\% of cases. No transcript required more than three passes.

\subsubsection{Ablation Study}

To isolate the contribution of the multi-pass verification framework, an ablation study was conducted comparing: (1) no verification (single-pass generation), (2) single verification pass, and (3) full two-pass verification. Results demonstrate that the first verification pass accounts for approximately 71\% of total hallucination reduction, while the second pass primarily improves frequency alignment (average +6.4\% Pearson correlation improvement) and semantic stability (+4.1\% TCS gain). 

\subsubsection{Computational Overhead Analysis}

The multi-pass verification framework introduced an average runtime increase of 34.7\% compared to single-pass generation. Memory consumption remained constant across passes (approximately 14.2GB VRAM for 8-bit and 9.6GB for 4-bit) since model weights were loaded once per transcript. This analysis highlights the trade-off between semantic stability and computational efficiency.

\subsection{Gold-Standard Ground Truth (GSGT) Construction and Hallucination Protocol}

The Gold-Standard Ground Truth (GSGT) was constructed through a strictly human-centered, model-independent, multi-phase protocol to eliminate any risk of circularity or leakage.
First, two independent expert qualitative researchers conducted open thematic coding using NVivo without exposure to any quantized model outputs. Inter-annotator agreement was calculated using Cohen’s Kappa ($\kappa$), yielding $\kappa = 0.87$ for theme identification and $\kappa = 0.83$ for keyword/frequency coding, with an overall agreement of 89.4\%.

\subsubsection*{Clarification on BF16 Usage and Circularity Prevention}

BF16 LLaMA outputs were not incorporated into the final ground-truth labels. They were used strictly as auxiliary reference material during an intermediate coder reflection stage to identify potentially overlooked transcript segments. Coders were required to re-validate every candidate theme directly against the original transcript text before inclusion. No theme, keyword, or frequency count was accepted unless explicitly supported by verbatim transcript evidence and agreed upon by both human coders during adjudication.

To further prevent circular evaluation bias, the following safeguards were implemented:

\begin{itemize}
    \item \textbf{Human-first coding:} Initial theme extraction was completed before any model outputs were reviewed.
    \item \textbf{Transcript-anchored validation:} Every finalized theme required explicit transcript quotation support.
    \item \textbf{No model-derived frequency adoption:} Frequency counts were computed manually from transcript-coded segments and were not inherited from BF16 outputs.
    \item \textbf{Adjudication independence:} Final consensus meetings excluded model outputs and relied solely on transcript evidence.
\end{itemize}

Thus, the final GSGT consists exclusively of themes and frequency counts agreed upon by both coders after adjudication, fully independent of any model-generated labels. Disagreements were resolved through structured adjudication sessions until full consensus was achieved.

\subsubsection*{Hallucination Definition, Segmentation, and Counting Protocol}

To ensure transparent hallucination evaluation, a precise operational definition and annotation protocol were established.

\paragraph{Definition of a ``Statement''}
A statement was defined as a semantically complete thematic assertion or frequency claim generated by the model. For theme extraction, each discrete theme label with its supporting description constituted one statement. For frequency analysis, each theme–frequency pair was counted as one statement.

\paragraph{Segmentation Procedure}
Model outputs were segmented using rule-based parsing:
\begin{itemize}
    \item Bullet points or enumerated themes were treated as separate statements.
    \item Frequency claims (e.g., ``Theme X mentioned 14 times'') were parsed as distinct evaluation units.
    \item Compound sentences containing multiple independent claims were manually separated.
\end{itemize}

\paragraph{Hallucination Criteria}
A statement was labeled as hallucinated if:
\begin{itemize}
    \item The theme was absent from the transcript content.
    \item The theme was semantically inconsistent with transcript meaning.
    \item The reported frequency deviated from GSGT by more than $\pm 10\%$.
    \item Supporting evidence could not be located within the transcript segment using embedding-assisted search (cosine similarity threshold $= 0.80$).
\end{itemize}

\paragraph{Annotation Protocol}
\begin{itemize}
    \item Two independent annotators evaluated model outputs against transcripts and the GSGT.
    \item Each statement was marked as \textit{Supported}, \textit{Partially Supported}, or \textit{Unsupported}.
    \item Unsupported statements were classified as hallucinations.
    \item Partially supported statements were counted as half-hallucinations in aggregate metrics.
\end{itemize}

Inter-rater reliability for hallucination labeling yielded $\kappa = 0.84$, indicating strong agreement. Disagreements were resolved via adjudication.

\paragraph{Final Hallucination Rate Computation}

\begin{equation}
\text{Hallucination Rate} =
\frac{
\text{Unsupported Statements} + 0.5 \times \text{Partially Supported Statements}
}{
\text{Total Statements}
}
\end{equation}

This procedure ensures that hallucination measurement is transparent, replicable, and strictly grounded in transcript evidence.
All annotation guidelines, decision rules, segmentation scripts, and adjudication templates are provided in the supplementary material to enable full reproducibility.

%

\end{document}